\documentclass[letterpaper, 10 pt, conference, twocolumn]{ieeeconf}  

\IEEEoverridecommandlockouts                              
\usepackage{ieeecref}
\usepackage{url}

\usepackage{amsmath}
\usepackage{amssymb}  
\usepackage[ruled,vlined,slide,commentsnumbered]{algorithm2e}
\usepackage{multirow}
\usepackage{array}

\usepackage{graphicx}
\usepackage{caption}
\usepackage{subcaption}

\usepackage{color}

\usepackage{color}

\newcommand{\be}{\begin{equation} \begin{aligned}}
\newcommand{\ee}{\end{aligned} \end{equation}}
\newcommand{\bea}{\begin{eqnarray}}
\newcommand{\eea}{\end{eqnarray}}


\title{\LARGE \bf
Towards Assistive Feeding with a General-Purpose Mobile Manipulator
}

\author{%
Daehyung Park*, You Keun Kim, Zackory M. Erickson, and Charles C. Kemp\\
\thanks{D. Park, Y. Kim, Z. Erickson, and C. C. Kemp are with Healthcare Robotics Lab, Georgia Institute of Technology, Atlanta, GA. *D. Park is the corresponding author \tt\small deric.park@gatech.edu.}%
}
\begin{document}

\maketitle
\thispagestyle{empty}
\pagestyle{empty}

\begin{abstract}
General-purpose mobile manipulators have the potential to serve as a versatile form of assistive technology. However, their complexity creates challenges, including the risk of being too difficult to use. We present a proof-of-concept robotic system for assistive feeding that consists of a Willow Garage PR2, a high-level web-based interface, and specialized autonomous behaviors for scooping and feeding yogurt. As a step towards use by people with disabilities, we evaluated our system with 5 able-bodied participants. All 5 successfully ate yogurt using the system and reported high rates of success for the system's autonomous behaviors. Also, Henry Evans, a person with severe quadriplegia, operated the system remotely to feed an able-bodied person. In general, people who operated the system reported that it was easy to use, including Henry. The feeding system also incorporates corrective actions designed to be triggered either autonomously or by the user. In an offline evaluation using data collected with the feeding system, a new version of our multimodal anomaly detection system outperformed prior versions.

\end{abstract}

\section{Introduction}
Activities of daily living (ADLs), such as feeding, toileting, and dressing, are important for quality of life \cite{wiener1990measuring}. Yet for many people with disabilities, such tasks prove challenging without assistance from a human caregiver. Numerous specialized assistive devices, including specialized robots, have been developed to help people with disabilities perform ADLs on their own. Each specialized device typically provides a narrow form of assistance suitable for people with particular impairments. In contrast, general-purpose mobile manipulators have the potential to provide assistance to diverse users with a wide variety of tasks \cite{chen2013robots}. Yet, the complexity of these general-purpose robots creates challenges, including the risk of being too difficult to use.

In this paper, we present a proof-of-concept robotic system for assistive feeding that makes use of a general-purpose mobile manipulator (a PR2 robot from Willow Garage). In our evaluation, people who operated the robot generally found the feeding system to be easy to use. The system also scooped and fed yogurt to the able-bodied participants with a high rate of success. In addition, we discuss an extended version of the progress-based anomaly detector that we introduced in \cite{park2016anomaly}. Our extended version classifies an anomaly using a support vector machine (SVM) with output from a hidden Markov model (HMM). In our evaluation, this new version outperformed prior versions.

\begin{figure}[t]
	\centering
    \includegraphics[trim={0cm, 1cm, 0cm, 0cm}, clip, width=55mm]{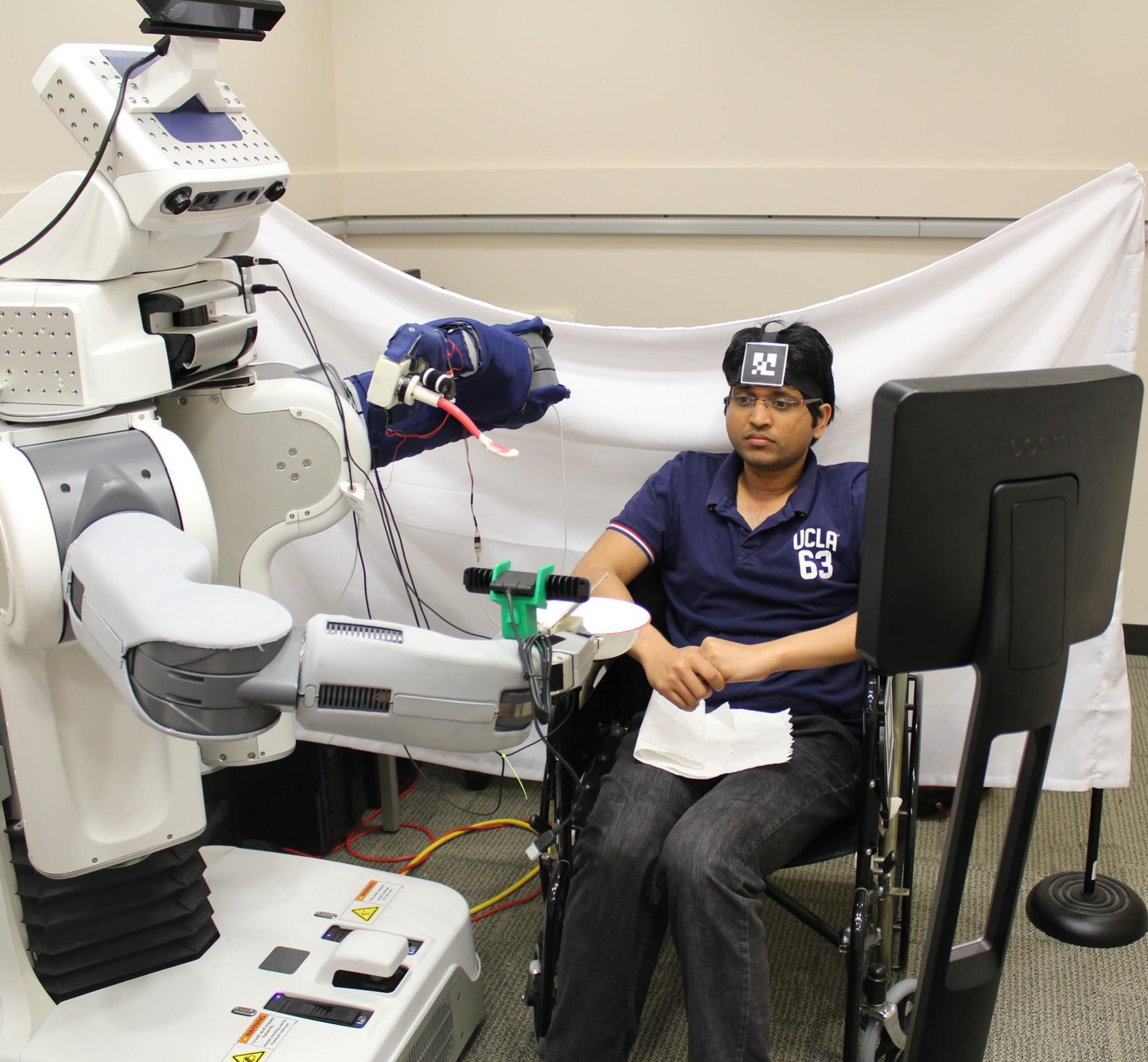}
    \caption{Our system feeding yogurt to an able-bodied person. A person with quadriplegia commanded the system remotely.}
	\label{fig: experiment}
\end{figure}

\section{Related Work}\label{sec:related_work}
Previous work has presented specialized robots that provide assistive manipulation for specific ADLs such as scratching, brushing, or feeding tasks \cite{ishii1995meal,topping2002overview}. Since the specialized design of these robots restricts their ability to generalize to other tasks, several studies have introduced general-purpose manipulators---such as 7-DoF arms mounted on a wheelchair or desk---to provide general assistance near the human, e.g. picking-and-placing \cite{hammel1989clinical, van1999provar, hillman1999wheelchair, hillman2002weston, dietsch2006portable, wakita2012user, kim2014system}. However, a fixed base restricts the scope of assistive tasks \cite{ari2015base}. Recently, several studies have introduced general-purpose mobile manipulators for various assistive robotic tasks, including shaving, picking-and-placing, and guiding tasks \cite{chen2013robots, hawkins2014assistive, ciocarlie2012mobile, graf2009robotic, nguyen2008assistive, kargov2005development}. To operate complex manipulators, researchers have also introduced easy-to-use control interfaces for users \cite{perez2015vibi, tsui2011want, tsui2007simplifying}.

Researchers have also investigated robotic feeding assistance \cite{topping2002overview, song2012novel, copilusi2015lab}. In addition to the absence of mobility, these previous robotic systems perform passive feeding actions that move food to a pre-designated location near the user, but require the user to actively catch the food, regardless of their disability. Takahashi and Suzukawa, on the other hand, introduced a feeding system with a human interface that allowed a user with quadriplegia to manually adjust feeding locations \cite{takahashi2006easy}. Similar to our work, Schröer et al. proposed an adaptive drinking assistance robot that finds the user's mouth with a vision system \cite{schroer2015autonomous}. However, their interface relies on electroencephalography (EEG).

\section{Overview of System Execution}
\begin{figure}[t]
	\centering
    \includegraphics[width=80mm]{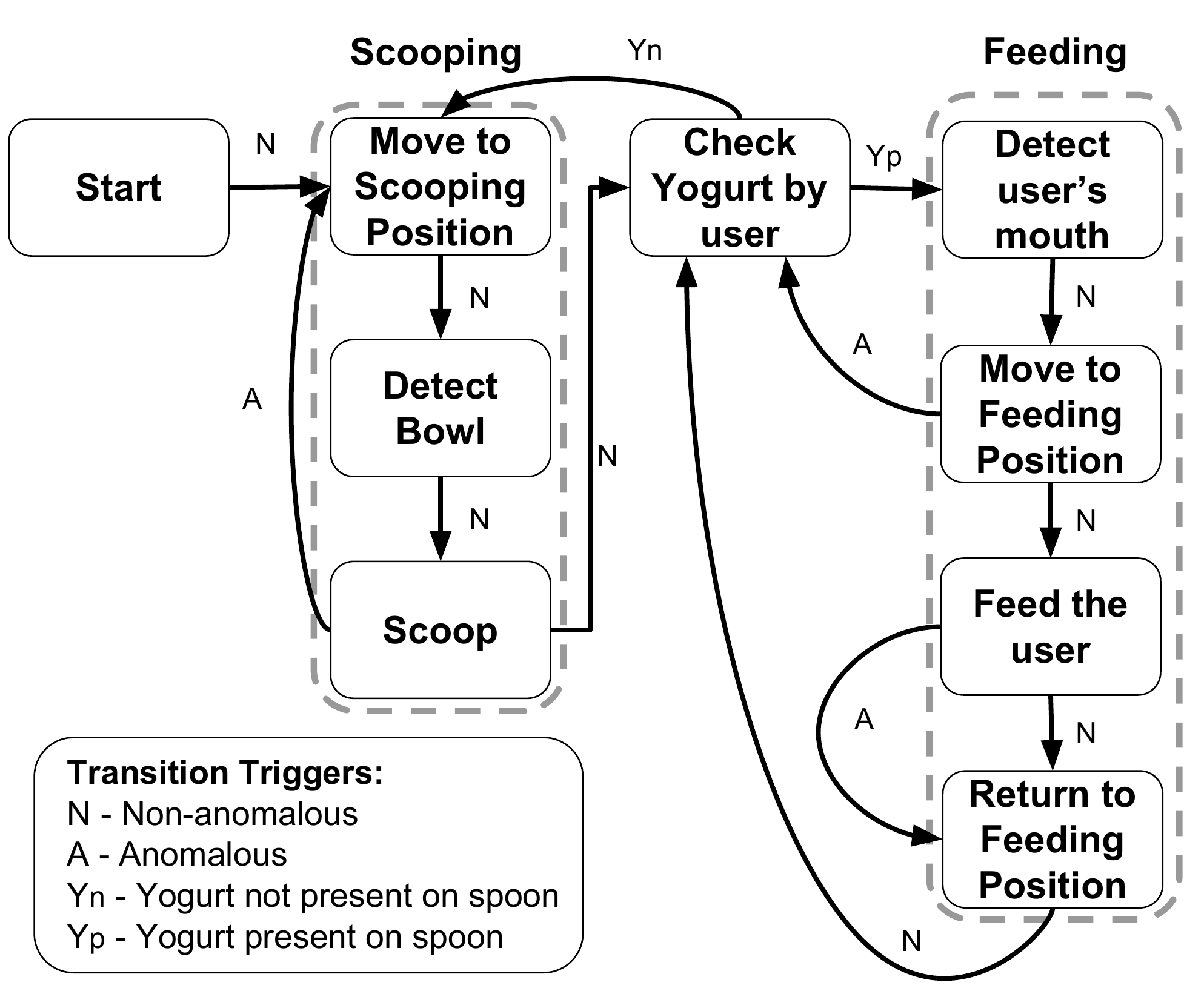}
	\caption{Scooping and feeding finite-state machine. Each Box represents a state, whereas arrows indicate state transitions.}
	\label{fig: motions}
\end{figure} 
%

\begin{figure}[t]
	\centering
	\includegraphics[trim={0cm, 0cm, 0cm, 0.3cm}, clip, width=7cm]{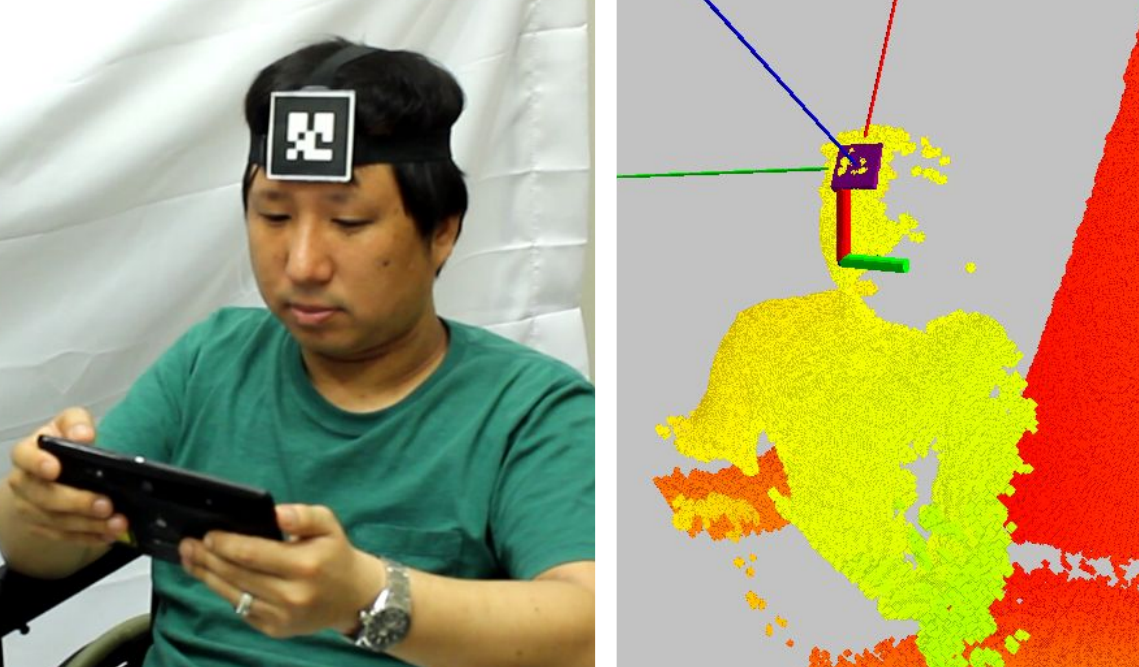}
	\caption{The robot estimate's the location of the user's mouth with a Kinect V2 and an ARTag attached to the user's head.}
	\label{fig: head_detection}
\end{figure}

%
Fig. \ref{fig: motions} shows the flow of the overall system using a finite-state machine (FSM). To perform the scooping task, the robot first estimates the location of the bowl and then scoops a spoon full of yogurt. The user can then provide feedback to the system on whether yogurt is present on the spoon. The robot re-executes the scooping task if no yogurt is present, otherwise the system proceeds to the feeding task. To perform the feeding task, the robot first determines a location for the user's mouth by detecting an ARTag attached to the user's head (as seen in Fig. \ref{fig: head_detection}). 
A user can send task commands to the PR2 via the web-based GUI, as well as stop or resume commands. The stop command is treated as an anomalous event and acts as specified in the FSM. After task completion, the user can provide feedback to further evaluate and tune the performance of the system.

\section{The System}

\subsection{Hardware}
\begin{figure}[t]
	\centering
    \includegraphics[width=80mm]{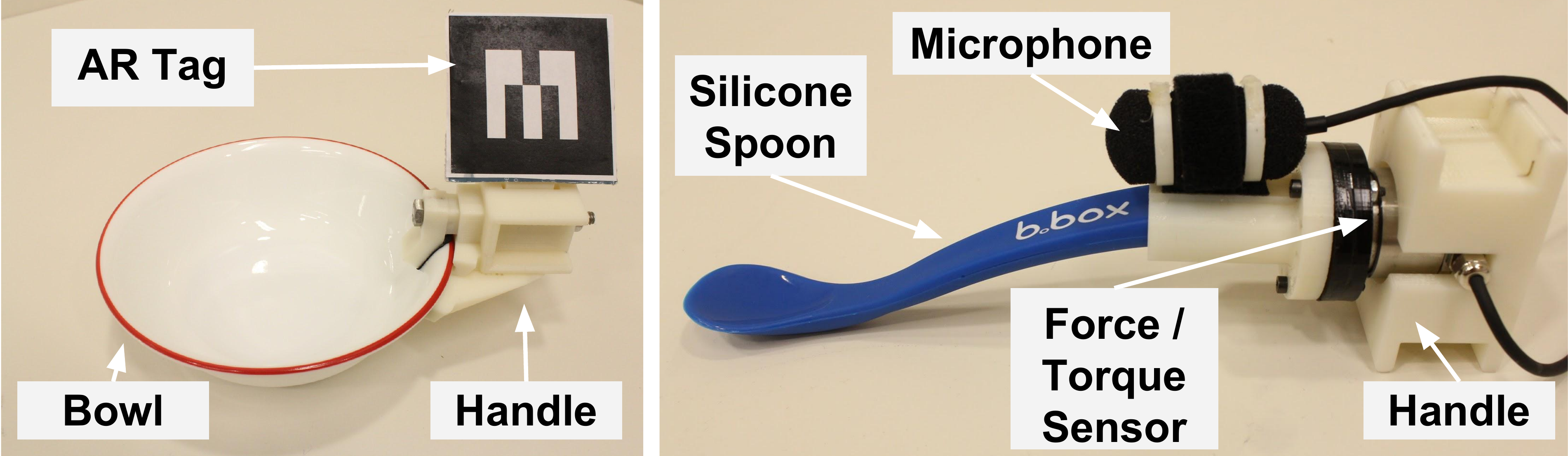}
    \caption{\textbf{Left:} The bowl for yogurt with an attached handle and ARTag. The PR2 can grip the handle and hold the bowl. \textbf{Right:} A tool for scooping and feeding that has a flexible silicone spoon, microphone, and force-torque sensor.}
	\label{fig: device}
\end{figure}

Our system uses a PR2 robot, a mobile manipulator from Willow Garage (see Fig. \ref{fig: experiment}). The PR2 consists of an omni-directional mobile base and two 7-DOF back-driverable arms with powered grippers. When moving the PR2's arms, our system uses low-gain impedance controllers at the joints. This is to reduce the likelihood of the robot applying high force to the person's body. The PR2's grippers firmly hold 3D printed handles, shown in Fig. \ref{fig: device}. On the other side of one handle, we attached a flexible silicone spoon, a directional microphone and a force-torque sensor for these scooping and feeding tasks. We also installed a Microsoft Kinect V2 RGB-D camera on the PR2's head to detect the person's mouth and track the location of the bowl. 

%

\subsection{Task Planning} \label{ssec: task_plan}
We incorporated four transition triggers to switch between states (as shown in Fig. \ref{fig: motions}). As expected, the multimodal anomaly detector described in Sec. \ref{ssec: anomaly_detection} can trigger an \textit{anomalous} transition. When an anomaly is detected, the system transitions to a state in which the robot halts or performs a corrective action. For instance, if a loud and unexpected sound is detected while feeding, an anomaly is triggered and the robot will retract its arm to avoid harming the user. Shared autonomy is also provided in which the user's input can trigger the $Y_p$ and $Y_n$ transitions. Following a scooping or feeding cycle, the user can confirm whether yogurt is present on the spoon. If there is not an adequate amount of yogurt on the spoon, the user can instruct the robot to re-scoop some yogurt. Finally, the user can provide feedback to the system after each task, which labels collected data to train the anomaly detector described in Sec. \ref{ssec: anomaly_detection}. 

\subsection{User Interface} \label{ssec: interface}
\begin{figure*}[ht]
	\centering	
	\includegraphics[trim={0.0cm, 0.7cm, 0cm, 0cm}, clip, height=4.0cm]{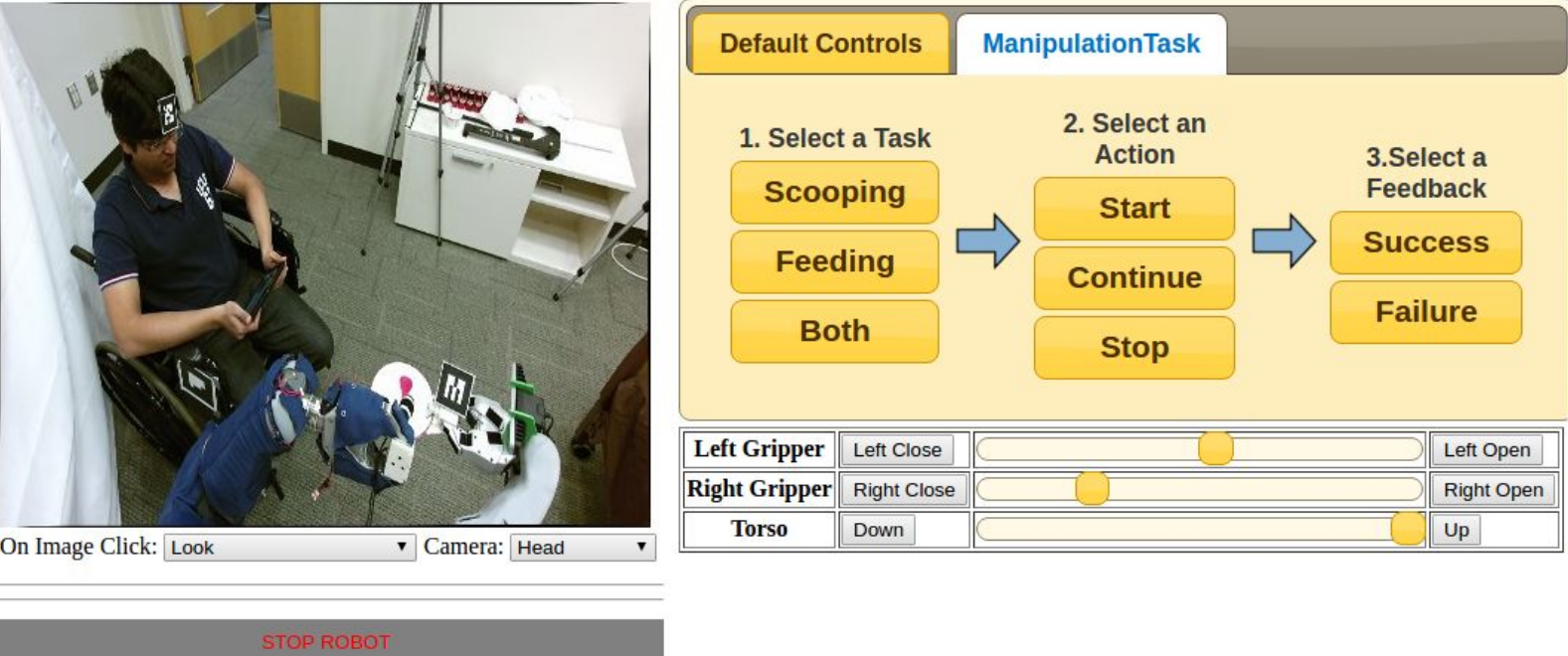}\hfill
	\includegraphics[trim={16.3cm, 0cm, 0cm, 0cm}, clip, height=40mm]{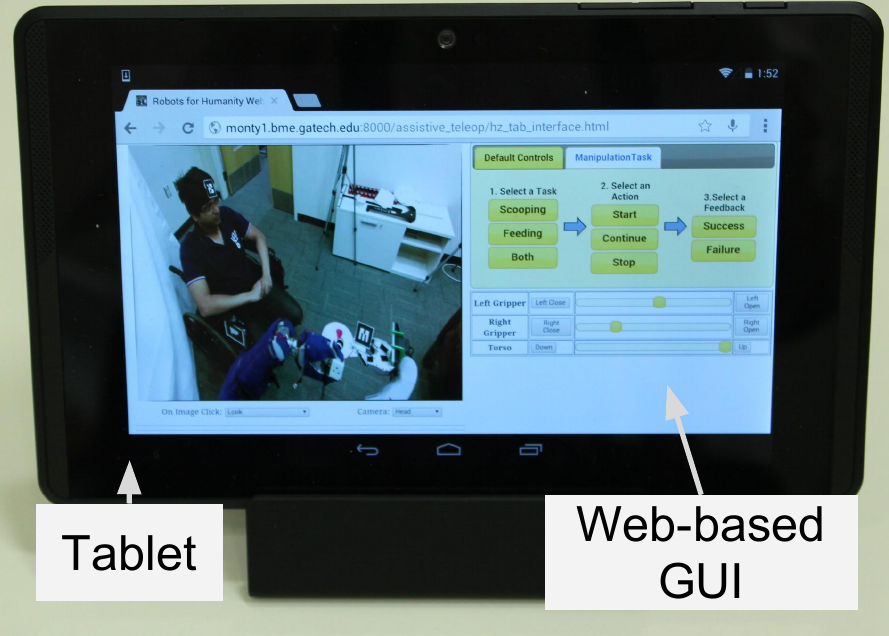}
	\caption{\textbf{Left:} The web-based GUI as described in Sec. \ref{ssec: interface}.
    \textbf{Right:} The GUI in a Chrome web-browser on a tablet.}
	\label{fig: gui}
\end{figure*}

We developed a web-based GUI to transmit task commands, display visual output from the Kinect, and collect feedback from users (see Fig. \ref{fig: gui}). In our evaluations, the users either operated the system with a tablet or a laptop web browser. The GUI consists of three sections; task selection, action selection, and evaluation. In the first section, users can select \textit{Scooping} or \textit{Feeding} motions. The selected motion can then be executed, resumed, or halted through the respective options in section 2 of the GUI. Note that the halting option triggers a corrective action described in Fig. \ref{fig: motions}. The user may then enter feedback (\textit{Success} or \textit{Failure}) after task completion. 

\subsection{Anomaly Detection } \label{ssec: anomaly_detection}
We extended the multimodal anomaly detector described in our previous work \cite{park2016anomaly}. The anomaly detector utilizes a multimodal HMM to model the sensory signals observed from non-anomalous operation of the system. Given non-anomalous training input, the HMM returns a vector of hidden state distributions, called execution progress, and a scalar called log-likelihood. Our previous version used a time-varying likelihood threshold (anomaly decision boundary) that came from clustering progress and log-likelihood pairs. 

\begin{figure}[t]
	\centering
	\includegraphics[width=8cm]{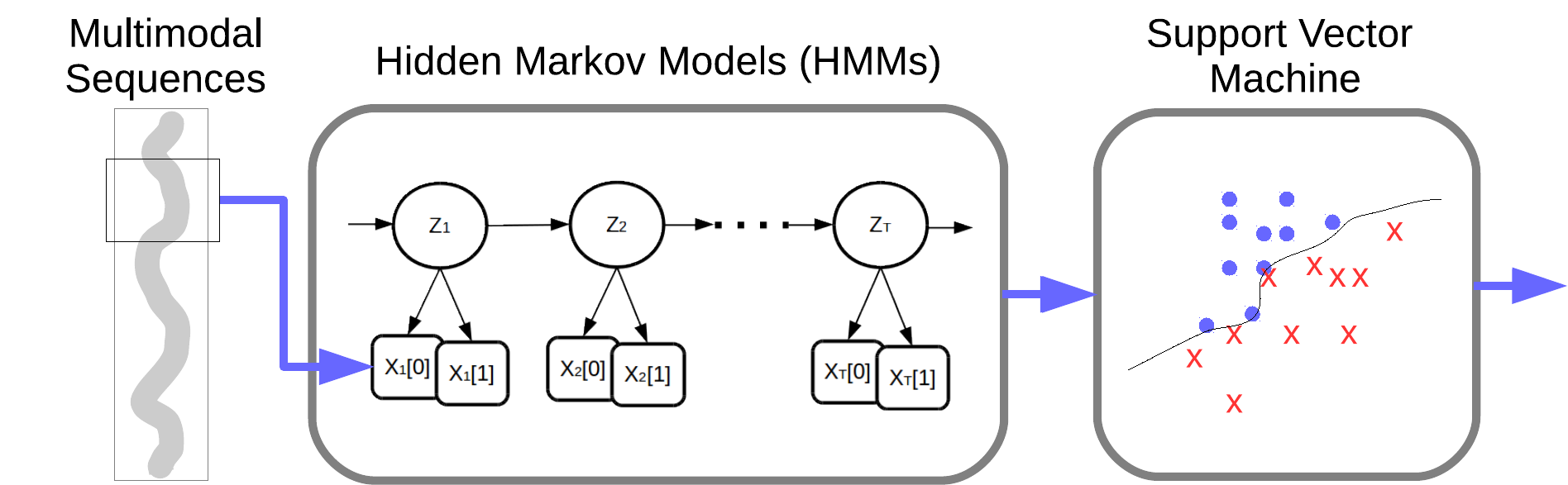}
    \caption{Anomaly detection framework: The HMM produces features from a multimodal observation sequence. Then, an SVM detects anomalies based on these features (Sec. \ref{ssec: anomaly_detection}).}
	\label{fig: anomaly_detector}
\end{figure}

In this work, we apply an SVM instead of clustering, which we refer to as an HMM-SVM anomaly detection approach. We provide the HMM-induced features, progress and likelihood, as the input to the SVM. Fig. \ref{fig: anomaly_detector} shows the sequence in which output from the HMM is processed by an SVM with a radial basis function kernel. The SVM then generates a nonlinear decision boundary for detecting anomalous observations. Furthermore, we can adjust the sensitivity of the anomaly detection method by unbalancing the class weights, $w_1$ and $w_{-1}$, of anomalous and non-anomalous observations. We varied the weight of anomalous observations, $w_1$, using the LIBSVM library \cite{CC01a}.   


\begin{figure*}[ht]
	\centering
	\includegraphics[width=17.5cm]{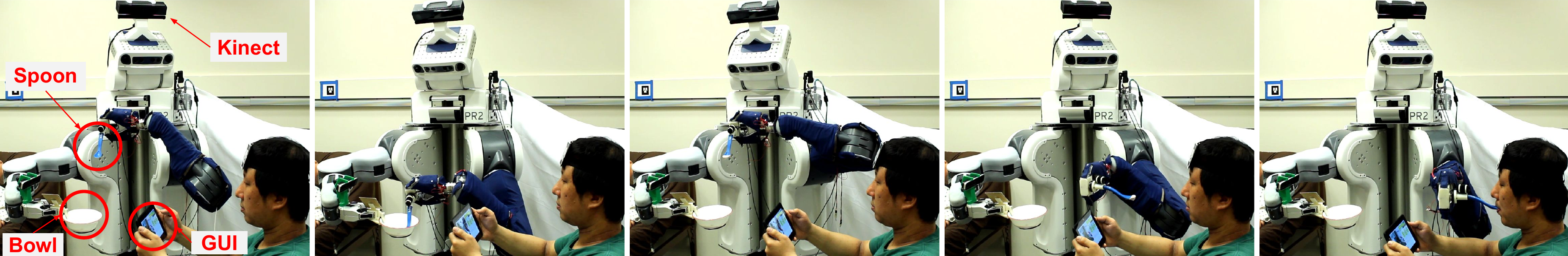}
	\caption{The images show the entire scooping and feeding process. The user operated the system via the GUI on the tablet.}
    
	\label{fig: scoop_feed_process}
\end{figure*} 

\section{Evaluation with 5 Able-bodied Participants}
We conducted a small evaluation of our system with approval from the Georgia Tech Institutional Review Board (IRB). As a step towards use by people with disabilities, we recruited 5 able-bodied participants. With the PR2 holding a bowl of yogurt, participants controlled the robot to scoop and feed themselves (see Fig. \ref{fig: scoop_feed_process}). We first briefly trained the 5 able-bodied participants to use the system. As part of this training, they practiced using it one or two times. Each practice run took about two minutes. Then, we asked them to freely use the scooping and feeding tasks. After each execution of a task, the participant labeled the execution as a \textit{success} or a \textit{failure} with the GUI (see \ref{fig: gui}). The participants used the scooping task 77 times and labeled 73 (94.8\%) of the executions as successful. They used the feeding task 83 times and labeled 82 (98.8\%) of the executions as successful. For the 5 failed executions, participants made statements such as `insufficient yogurt on the spoon,' `spilled yogurt from the spoon,' and `missed timing of closing mouth.' 


We then asked the participants to fill out a survey with 13 questions (five-point Likert items) based on \cite{chen2013robots}. The following table provides the questionnaire results that we found most informative. The columns provide response counts for strongly disagree (sd), disagree (d), neither (n), agree (a), and strongly agree (sa). The plus sign (+) denotes Henry Evans's responses. Overall, participants found the system to be easy to use and felt safe using it. However, their responses suggest that the comfort and speed of the system ($\sim$2 minutes per scoop fed) could be improved.

\begin{center}
{\footnotesize
\setlength{\tabcolsep}{2pt}
\begin{tabular}{lccccc}
question     	         & sd & d & n & a & sa \\
\hline
The system was easy and intuitive to use & 0 & 0 & 1 & 1 & 3+ \\
The system was comfortable to use & 0 & 2 & 0 & 2 & 1 \\
The web interface layout and icons were intuitive & 0 & 0 & 0 & 3 & 2+ \\
I felt safe during the experiment & 0 & 0 & 0 & 1 & 4 \\
I was satisfied with the time it took to complete the task & 0 & 2+ & 1 & 1 & 1 \\
\end{tabular}
}
\end{center}

\begin{figure}[t]
	\centering
    \begin{subfigure}[b]{0.5\columnwidth}
    	\centering
		\includegraphics[trim=0.2cm 0.2cm 0.2cm 0.2cm, clip=true, width =4cm]{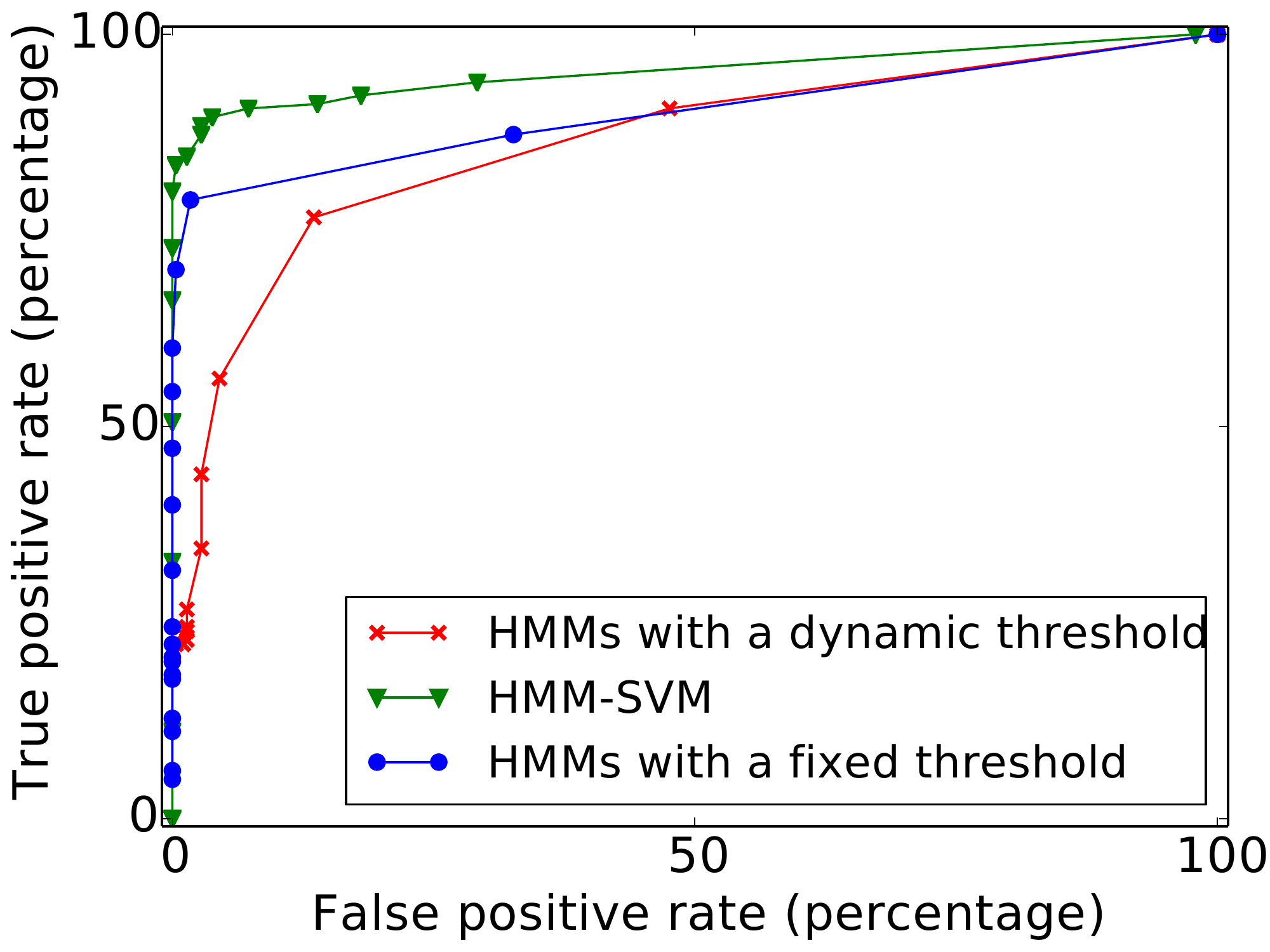}
        \caption{Scooping}
    \end{subfigure}%
    \begin{subfigure}[b]{0.5\columnwidth}
    	\centering
		\includegraphics[trim=0.2cm 0.2cm 0.2cm 0.2cm, clip=true, width=4cm]{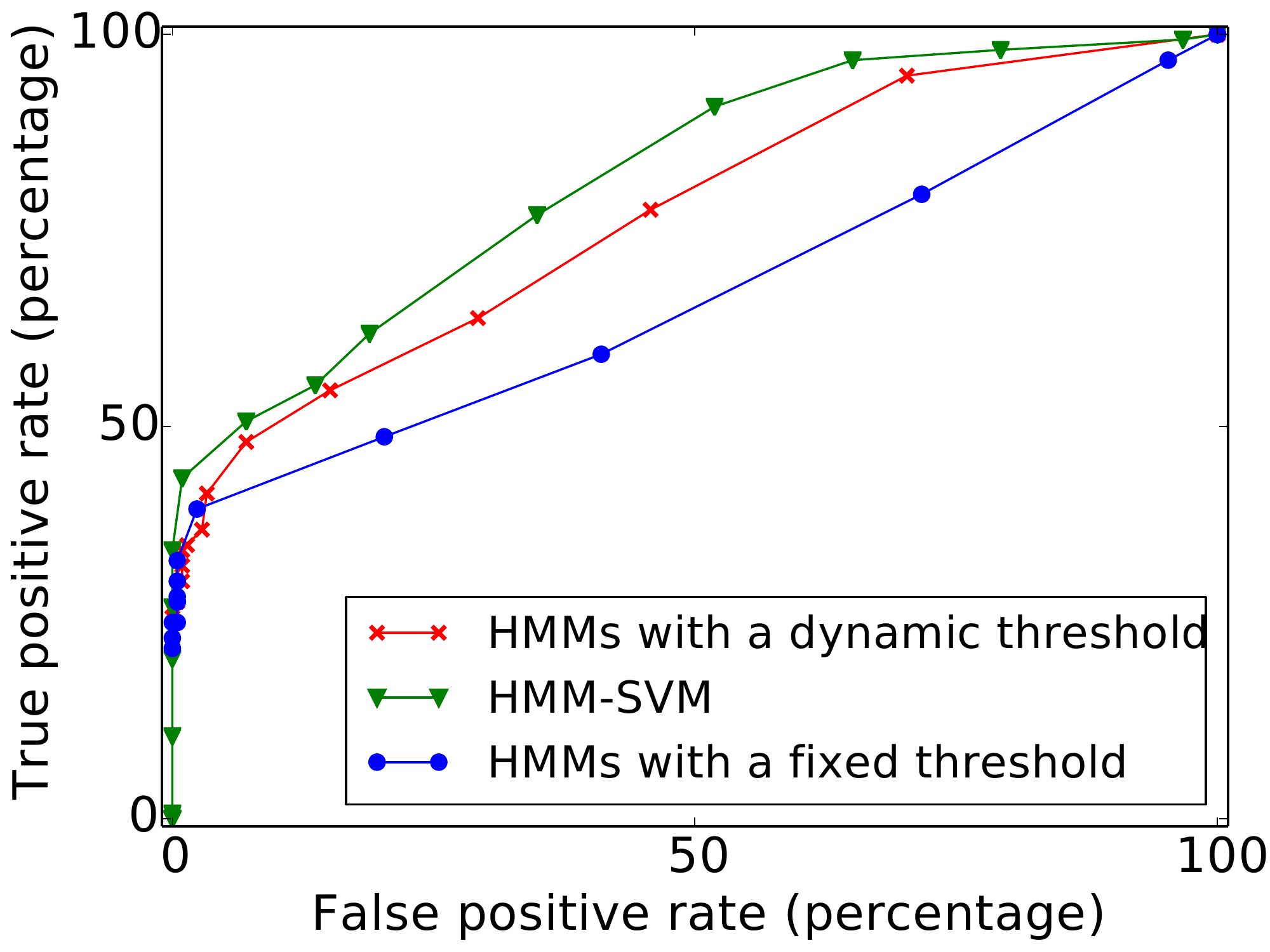}
        \caption{Feeding}        
    \end{subfigure}%
	\caption{ Receiver operating characteristic (ROC) curves. }
	\label{fig: ROC}
\end{figure}

To evaluate our HMM-SVM anomaly detector, we also asked each participant to perform 10 intentionally anomalous actions, such as pushing the PR2's arm while it was scooping yogurt, making loud or unexpected sounds, or refusing to open their mouths during feeding. After collecting this data, we removed 14 outliers by hand to avoid convergence issues during HMM training. We used the remaining 72 non-anomalous and 86 anomalous iterations for the scooping task, and 53 non-anomalous and 39 anomalous iterations for the feeding task. Fig. \ref{fig: ROC} displays ROC curves that evaluate the performance of the detector with respect to the false positive rate and true positive rate. We performed 4-fold cross validation varying the class weight of anomalous data in the SVM. For both tasks, our HMM-SVM approach outperformed other HMM anomaly detectors that utilized fixed or dynamic threshold methods. 


\section{Remote Evaluation by Henry Evans}
%





We also performed a test with Henry Evans, a person with severe quadriplegia with whom our lab collaborates with approval from the Georgia Tech Institutional Review Board (IRB). From his home in California, USA, he used the web-based GUI to command a PR2 robot in Georgia, USA to feed an able-bodied person. Henry used an off-the-shelf head tracker and a mouse button to operate the web-based GUI. While using the system, he had visual feedback from the web-based GUI and a Beam+ (a separate telepresence robot). Henry successfully used the system to feed yogurt to the able-bodied participant. When applicable, Henry answered the survey questions given to the able-bodied participants. His answers, shown in the previous table with a plus sign (+), were similar to the able-bodied participants'. In response to an open-ended question at the end of the survey, he wrote ``overall, worked well, although the PR2 video did not work." In a later email with the subject line "feeding feedback", he wrote "it is ready for field testing!", indicating he is prepared to try out the system in person.



\section{Conclusion}
We introduced an assistive feeding system that employs a general-purpose mobile manipulator (a PR2 robot). The system provides a high-level web-based interface to the operator that people generally found easy to use. The system successfully fed yogurt to able-bodied participants and they felt safe with the system. In offline tests, our new method of anomaly detection outperformed our previous methods. Overall, our results suggest that it is feasible for general-purpose mobile manipulators to provide feeding assistance, although further work remains. In particular, people with disabilities have not eaten with our current system.




\textit{\textbf{Acknowledgment}: This work was supported in part by NSF Awards IIS-1150157 and NIDILRR grant 90RE5016-01-00 via RERC TechSAge.}

\begin{small}
\bibliographystyle{ieeetr}
\bibliography{humanoids}
\end{small}

\end{document}